\documentclass[letterpaper]{article} % DO NOT CHANGE THIS
\usepackage[preprint]{aaai2027}  % DO NOT CHANGE THIS
\usepackage[hyphens]{url}  % DO NOT CHANGE THIS
\usepackage{graphicx} % DO NOT CHANGE THIS
\usepackage{natbib}  % DO NOT CHANGE THIS AND DO NOT ADD ANY OPTIONS TO IT
\usepackage{caption} % DO NOT CHANGE THIS AND DO NOT ADD ANY OPTIONS TO IT
\usepackage{algorithm}
\usepackage{algorithmic}
\usepackage{amsmath}
\usepackage{amssymb}

\usepackage{newfloat}
\usepackage{listings}
\DeclareCaptionStyle{ruled}{labelfont=normalfont,labelsep=colon,strut=off} % DO NOT CHANGE THIS
\floatstyle{ruled}
\newfloat{listing}{tb}{lst}{}
\floatname{listing}{Listing}

\usepackage{booktabs}

\title{SearchMaster: Grounded and Regulated Self-Play for Search Agents}
 \author{
      Wentao Tan\textsuperscript{\rm 1},
      Qiong Cao\textsuperscript{\rm 1}\textsuperscript{\dag},
      Jiaqi Wang\textsuperscript{\rm 1},
      Nan Duan\textsuperscript{\rm 1}
  }
  \affiliations{
      \textsuperscript{\rm 1}JD Future Academy, Beijing\\
      \textsuperscript{\dag}Corresponding author.\\
      \{tanwentao1, caoqiong1\}@jd.com
  }

\begin{document}

\maketitle

\begin{abstract}
Training LLM-based search agents requires high-quality search data: tasks that demand genuine multi-hop retrieval and trajectories that use search tools effectively. Existing pipelines often depend on human-written tasks, expert demonstrations, or stronger teacher models. We present SearchMaster, a self-play framework that trains a shared LLM policy from search tasks it generates, solves, and verifies in a local search environment. The key challenge is that self-generated tasks and rollouts can yield misleading signals: pseudo multi-hop questions, success-rate difficulty estimates that ignore search depth, and rollouts with excessive opening but little targeted evidence acquisition. SearchMaster addresses these failure modes with three controls. An Evidence-Chain Generator (ECG) grounds task generation in explicit cross-document evidence chains to reduce pseudo multi-hop questions. A Search-Depth Reward (SDR) scores task difficulty by the search depth of successful rollouts rather than success rate alone, keeping retained tasks search-intensive. An Over-Opening Penalty (OOP) regulates tool use by discouraging excessive document opening, avoiding long but shallow browsing. Verified Proposer and Solver rollouts are then jointly optimized with GRPO. Across six deep-search benchmarks, SearchMaster raises average accuracy from 38.19\% to 51.52\% with a Qwen3.5-9B backbone and from 51.25\% to 62.54\% with a Qwen3.5-27B backbone. On BrowseComp-Plus, the corresponding gains are 30.1 and 17.8 percentage points. These results show that grounded and regulated self-play provides effective search-agent training data across model scales, without human-labeled QA pairs or expert demonstrations. The code is available at \url{https://github.com/WentaoTan/SearchMaster}.

\end{abstract}

\begin{figure}[h]
  \centering
  \includegraphics[width=\linewidth]{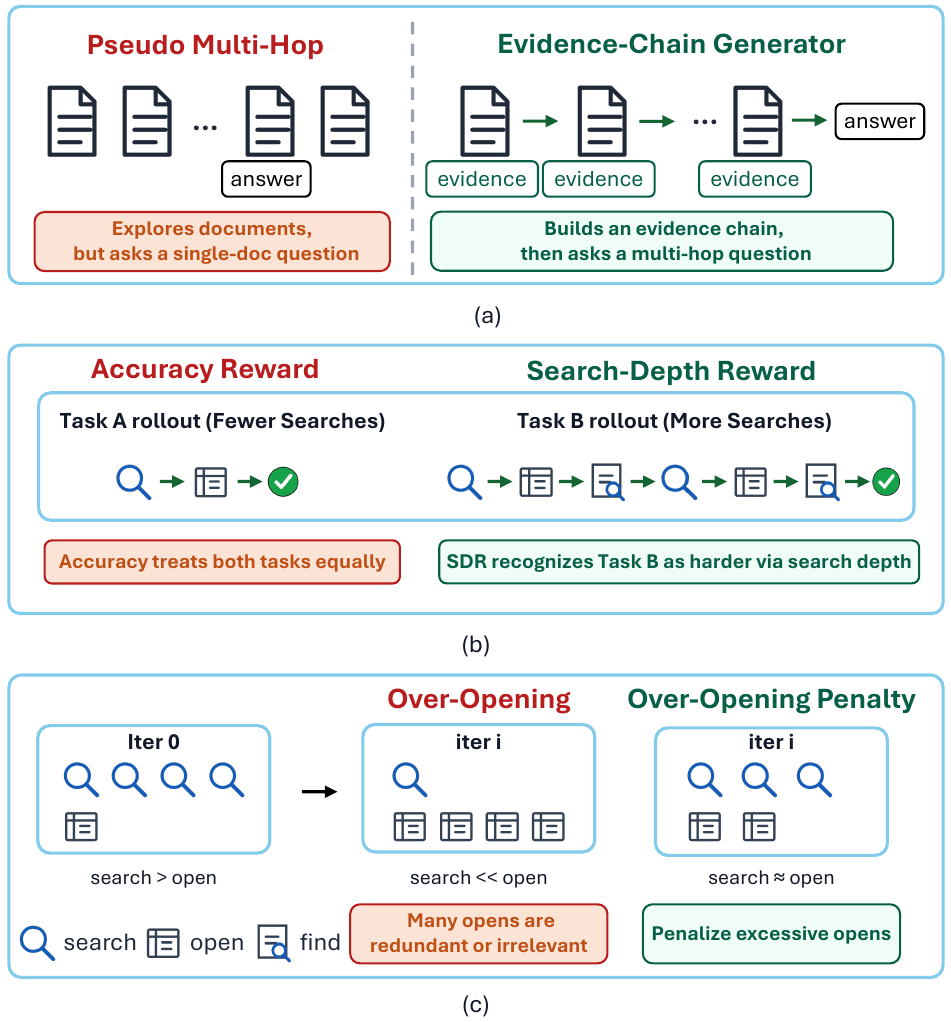}
  \caption{SearchMaster grounds tasks and regulates behavior. (a) ECG builds explicit evidence chains to reduce pseudo multi-hop questions. (b) SDR rewards deeper successful rollouts, favoring multi-step search. (c) OOP penalizes excessive document opening to reduce long but shallow browsing.}
  \label{fig:searchmaster-overview}
\end{figure}

\section{Introduction}

LLM-based search agents improve knowledge-intensive question answering by retrieving evidence from the web~\cite{nakano2021webgpt,yao2022react,li2026webthinker,wei2025webagent}. Recent work has improved this capability either through reinforcement learning for tool use~\cite{jin2025search,song2025r1,zhao2026r,wang2025stepsearch,shi2026search} or by constructing multi-turn search data for supervised training~\cite{zheng2025deepresearcher,li2026openresearcher,chu2026redsearcher,li2025websailor}. Despite different training objectives, both rely on high-quality search data: questions that require cross-document retrieval and trajectories that acquire evidence through targeted search. Obtaining such data is expensive, often requiring human annotation or expert demonstrations.
  
Self-play appears to be a natural solution: the model can generate search tasks and learn by solving them. Existing search self-play methods have explored this direction by constructing questions from predefined answers~\cite{lu2025search} or through open-ended exploration~\cite{yue2026dr}. While these approaches reduce reliance on externally written tasks, search-agent self-play has a distinct failure mode: self-generated tasks and rollouts can look useful while providing weak or misleading supervision. For example, the model may browse multiple documents yet still write a question answerable from one of them alone. Moreover, prior search self-play methods typically proxy task difficulty by rollout accuracy, often favoring tasks with moderate success rates; however, this signal may fail to distinguish shallow lookup tasks from those requiring multi-step search.  Finally, document opening often exposes useful evidence, but the model may gradually over-rely on it, revisiting already-seen documents instead of acquiring targeted evidence. Thus, the central challenge is not merely generating more data, but stabilizing the self-play loop so that its training signals reinforce evidence-grounded tasks, multi-step search, and efficient tool use.

To address this challenge, we propose \textbf{SearchMaster}, a grounded and regulated self-play framework. SearchMaster uses a shared trainable policy for two roles: a \textbf{Proposer}, which generates multi-hop search tasks within a local search environment, and a \textbf{Solver}, which answers these tasks through browser-tool rollouts. A frozen \textbf{Verifier} evaluates task quality and solution accuracy, turning verified rollouts into rewards for GRPO training~\cite{shao2024deepseekmath}. SearchMaster operationalizes this grounded-and-regulated principle through three mechanisms that jointly ground self-generated tasks and regulate the search behavior they induce (Figure~\ref{fig:searchmaster-overview}). The \textbf{Evidence-Chain Generator (ECG)} requires the Proposer to incrementally build an explicit cross-document evidence chain during exploration and derive the question from the full chain. The \textbf{Search-Depth Reward (SDR)} scores a task by the search depth of its successful rollouts, encouraging the Proposer to generate tasks that require deeper retrieval. The \textbf{Over-Opening Penalty (OOP)} constrains the open-to-search ratio, discouraging redundant opening without penalizing tasks that legitimately require inspecting multiple documents. 

Overall, our main contribution is to formulate search-agent self-play as a problem of grounding and regulating self-generated supervision. This perspective turns a potentially noisy self-play loop into one that promotes cross-document task grounding, search-depth-aware task selection, and balanced tool use. Across six deep-search benchmarks, SearchMaster improves the Qwen3.5-9B backbone \cite{qwen3.5} from 38.19\% to 51.52\% average accuracy (\(+13.3\) points) and the Qwen3.5-27B backbone \cite{qwen3.5} from 51.25\% to 62.54\% (\(+11.3\) points). On BrowseComp-Plus \cite{chen2026browsecomp}, the respective gains are 30.1 points (30.12\% to 60.24\%) and 17.8 points (54.46\% to 72.29\%). The consistent improvements at both scales show that the method is not tied to a particular model size or initial capability. Notably, these gains rely solely on a local search environment, without human-labeled data or expert supervision. We will release the code, trained checkpoints, and self-generated training data.

\section{Related Work}

\subsection{Search Agent Training}

Early search agents primarily relied on prompting or human demonstrations for web tool use \cite{guu2020retrieval,lewis2020retrieval,schick2023toolformer}. For instance, WebGPT \cite{nakano2021webgpt} trained language models to browse and cite web evidence using human demonstrations, while ReAct \cite{yao2022react} used prompting to interleave reasoning with tool usage. These approaches show that external evidence improves answer reliability; however, achieving stable long-horizon search behavior demands further training.

Recent efforts to improve search-agent training broadly fall into two categories. The first focuses on optimizing search behavior through reinforcement learning and reward design on existing QA tasks \cite{tan2026rag,mei20252}. Search-R1 \cite{jin2025search} and R1-Searcher \cite{song2025r1} use outcome-based rewards to encourage search during reasoning. Recognizing the limitations of final-answer rewards in guiding long trajectories, subsequent methods incorporate richer process-level signals: R-Search \cite{zhao2026r} introduces multiple rewards for search and reasoning quality, StepSearch \cite{wang2025stepsearch} applies step-wise feedback to improve retrieval and reduce redundancy, AutoRefine \cite{shi2026search} iteratively refines retrieved knowledge across search calls, and AutoSearch \cite{sun2026autosearch} adapts search depth to balance accuracy and efficiency.

The second line of work aims to construct higher-quality training data by generating multi-turn search trajectories and challenging tasks. OpenResearcher \cite{li2026openresearcher} builds an offline search environment and leverages a teacher model to produce multi-turn search-and-browse trajectories. WebSailor \cite{li2025websailor} emphasizes task difficulty by creating high-uncertainty web tasks through structured sampling and information obfuscation. Extensions include synthetic data generation, multi-agent collaboration, and staged post-training \cite{chu2026redsearcher,liu2025webexplorer,yao2026researcher}. Despite these advances, acquiring search-worthy tasks and trajectories remains costly, often relying on human annotation, expert models, or complex generation pipelines.

Overall, while these approaches advance search-agent training, obtaining scalable multi-hop search data remains a challenge. This motivates self-play, where models generate and learn from their own search experiences.

\subsection{Self-Play for Search Agent Training}

Self-play offers a promising avenue to reduce reliance on external labeled data. In reasoning and tool-use domains, prior work shows that models can bootstrap learning by generating tasks and solutions themselves when validity and correctness can be verified \cite{huang2025r,zhao2026absolute,acikgoz2026tool}. However, self-play for search agents is more challenging, since generated tasks must be grounded in evidence, require genuine multi-hop search, and induce meaningful rather than shallow tool-use trajectories.

Search Self-Play \cite{lu2025search} is an early attempt at self-play for search-agent training: it samples a target answer from a predefined set, and the Proposer searches for supporting evidence to formulate a question pointing to it. While this enables self-play, task generation remains constrained by predefined answers rather than open-ended exploration. Dr.~Zero \cite{yue2026dr} instead lets the Proposer generate questions directly through external search, rewarding task difficulty by the Solver's success rate. This moves toward more autonomous task generation, but success rate remains a coarse signal: it reflects whether the Solver answers correctly, not whether the task requires nontrivial search beyond a shallow lookup. Beyond this difficulty-estimation issue, pseudo multi-hop questions and over-opening drift remain underexplored. SearchMaster addresses these search-specific issues by making evidence construction, task difficulty, and tool-use balance explicit parts of the self-play loop, enabling more reliable self-evolution of search agents.

\begin{figure*}[ht]
  \centering
  \includegraphics[width=\textwidth]{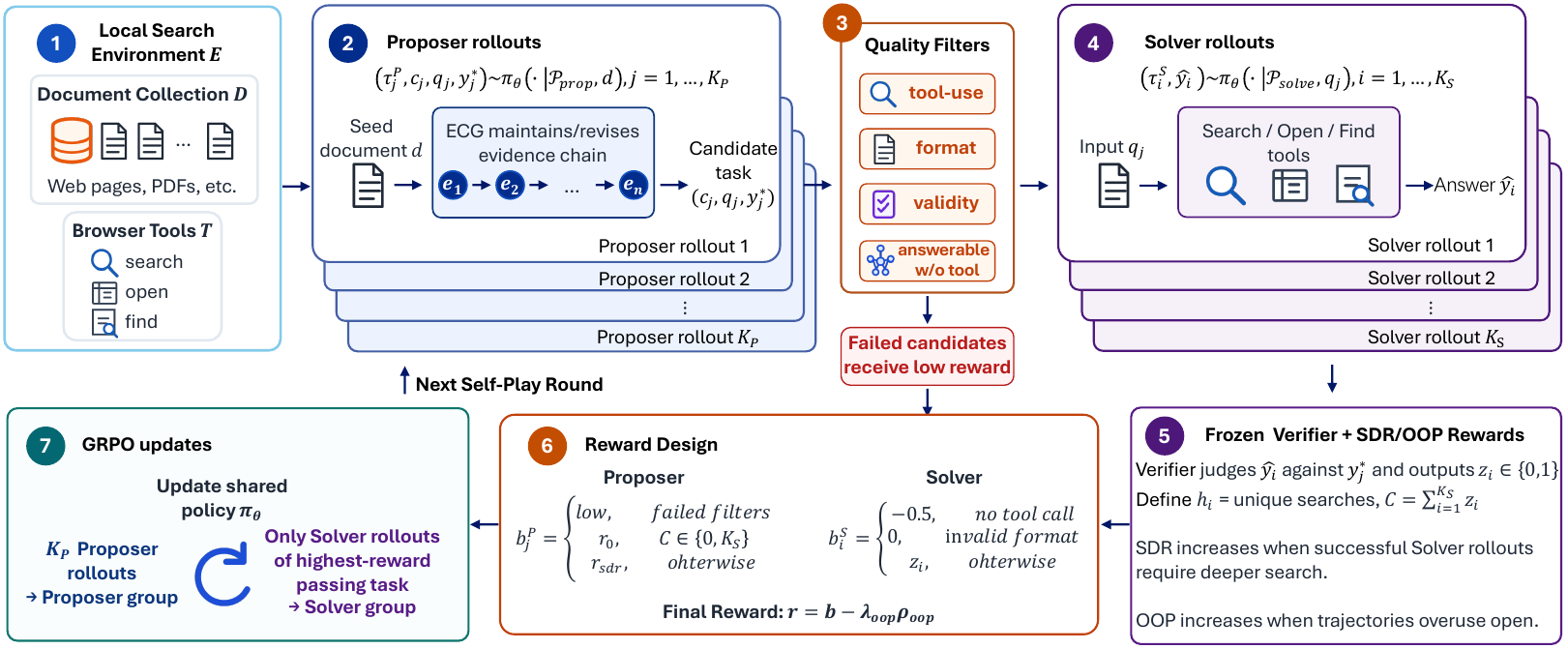}
  \caption{Overview of the SearchMaster self-play loop. A shared policy \(\pi_\theta\) acts as both Proposer and Solver in a local search environment. The Proposer builds evidence-chain tasks (ECG), filters remove invalid candidates, and the Solver answers passing tasks through independent rollouts. A frozen Verifier, SDR, and OOP provide rewards, and verified Proposer/Solver rollouts are optimized jointly with GRPO.}
  \label{fig:searchmaster-framework}
\end{figure*}

\section{Method}
\label{sec:method}

\subsection{Problem Setup and Framework Overview}

Figure~\ref{fig:searchmaster-framework} illustrates the SearchMaster framework. It trains a single policy \(\pi_\theta\) within a search environment \(\mathcal{E}\) constructed over document collection \(\mathcal{D}\). The policy interacts with \(\mathcal{E}\) through browser tools \(\mathcal{T} = \{\texttt{search}, \texttt{open}, \texttt{find}\}\), where \texttt{search} returns ranked documents for a query, \texttt{open} reveals a document's content, and \texttt{find} locates exact matches within opened documents.

During training, the same policy \(\pi_\theta\) acts as both the \textbf{Proposer} \(P_\theta\) and the \textbf{Solver} \(S_\theta\). The \textbf{Verifier} \(V\) is a frozen copy of the initial model and remains fixed.

Each self-play iteration samples unlabeled seed documents \(d\) from \(\mathcal{D}\). For each seed, the Proposer samples \(K_P\) task-generation rollouts, each maintaining an ECG evidence chain \(c\) and producing a candidate task \((c,q,y^\star)\). Quality filters remove invalid candidates before Solver rollouts.
For each passing task, the Solver receives only \(q\) and produces \(K_S\) answer rollouts. The frozen Verifier judges each answer, producing correctness scores \(z_i\), and we record the search depth \(h_i\) of each rollout as the number of unique search queries. These signals define rewards for both roles: \(z_i\) rewards Solver correctness, successful-rollout depths estimate task difficulty for the Proposer through SDR, and OOP is subtracted from both rewards to discourage excessive opening. The resulting samples form a Proposer GRPO group over the \(K_P\) task-generation rollouts and a Solver GRPO group over the \(K_S\) answer rollouts of the highest-reward passing task. Since one seed can produce multiple passing tasks and up to \(K_P \times K_S\) Solver rollouts, using all Solver rollouts would make Solver samples dominate the update; we therefore keep only the \(K_S\) Solver rollouts of the highest-reward passing task.

\subsection{Evidence-Chain Task Generation}

Generating genuine multi-hop search tasks is challenging because pseudo multi-hop tasks can arise even when the Proposer is explicitly instructed to generate questions requiring cross-document evidence. In one failure mode, the Proposer stays close to the seed, searching for surface-level facts and asking a question answerable from the seed alone. In another, it reaches related documents but does not encode the cross-document links into the question, leaving the final answer recoverable from just one visited document. Examples of both cases are provided in the Supplementary Material.

To address this, the Evidence-Chain Generator (ECG) turns task proposal into explicit chain construction. Starting from a seed document \(d\), the Proposer repeatedly follows entity or fact links across documents: it selects a salient entity, searches for a connected document, identifies a new entity or fact, and extends the chain. The question is then generated with explicit requirements for full-chain necessity and no single-document answer.

Formally, for each \(d\), the Proposer samples \(K_P\) independent rollouts:
\begin{equation}
(\tau_j^P, c_j, q_j, y_j^\star)\sim\pi_\theta(\cdot\mid\mathcal{P}_{\mathrm{prop}}, d),\qquad j=1,\dots,K_P,
\label{eq:proposer-rollout}
\end{equation}
where \(\mathcal{P}_{\mathrm{prop}}\) is the Proposer prompt (Supplementary Material), \(\tau_j^P\) the tool-use trace, \(c_j = e_{j,1} \rightarrow e_{j,2} \rightarrow \cdots \rightarrow e_{j,n_j}\) the evidence chain of entities or facts, \(q_j\) the question generated from \(c_j\), and \(y_j^\star\) its reference answer. By deriving \(q_j\) from this constrained chain rather than a single document, ECG biases task generation toward questions that require combining evidence across documents.

Before launching Solver rollouts, SearchMaster applies four filters to each candidate task:
\begin{itemize}
    \item \textbf{Tool-use filter:} treats candidates with too few \texttt{search} or \texttt{open} actions as shallow.
    \item \textbf{Format filter:} checks that \((c_j, q_j, y_j^\star)\) are present and that \(c_j\) contains at least two evidence items.
    \item \textbf{Validity filter:} asks the Verifier whether \(q_j\) is valid, \(y_j^\star\) is supported by \(\tau_j^P\), and the chain is logically sound.
    \item \textbf{Parametric-knowledge filter:} removes questions the Verifier can answer without tools.
\end{itemize}
A candidate that fails any filter receives a low Proposer reward, whereas those passing all four are sent to the Solver.

\subsection{Solver Rollouts and Search Depth}

After a task passes the filters, the Solver receives only the question \(q_j\) and produces \(K_S\) independent answer rollouts
\begin{equation}
(\tau_i^S,\hat{y}_i)\sim\pi_\theta(\cdot\mid\mathcal{P}_{\mathrm{solve}},q_j),\qquad i=1,\dots,K_S,
\label{eq:solver-rollout}
\end{equation}
where \(\mathcal{P}_{\mathrm{solve}}\) is the Solver prompt (Supplementary Material), \(\tau_i^S\) is the tool-use trace, and \(\hat{y}_i\) is the final answer. The frozen Verifier compares \(\hat{y}_i\) with the reference answer \(y_j^\star\) and outputs a correctness score \(z_i\in\{0,1\}\). We also compute the search depth \(h_i\) as the number of unique search queries in \(\tau_i^S\), and let \(C=\sum_{i=1}^{K_S}z_i\) denote the number of successful rollouts. These signals play different roles: \(z_i\) rewards Solver correctness, while the search depths of successful rollouts estimate how much search the proposed task requires for the Proposer reward.

\subsection{Reward Design and Over-Opening Penalty}

The Proposer reward favors tasks that are solvable but still challenging for the current Solver. Candidates that fail the quality filters receive low rewards. Passing tasks with \(C=0\) or \(C=K_S\) receive only a low base reward \(r_0=0.2\), since they are respectively too hard or too easy. If all \(K_P\) tasks generated from a seed receive rewards that do not exceed \(r_0\), we discard all data generated from that seed.

For the remaining tasks with \(0<C<K_S\), success rate alone does not distinguish shallow lookup from deeper search. We therefore measure task difficulty by the easiest successful search depth,
\begin{equation}
h_{\min}=\min_{i:z_i=1}h_i,
\label{eq:hmin}
\end{equation}
rather than by the mean or maximum search depth. If any successful rollout solves the task with shallow search, the task admits a shortcut and should not receive a high difficulty reward. We then compute the Search-Depth Reward (SDR):
\begin{equation}
r_{\mathrm{sdr}}=r_0+(1-r_0)\min\!\left(\frac{h_{\min}}{H_{\mathrm{sdr}}},1\right),
\label{eq:sdr}
\end{equation}
where \(H_{\mathrm{sdr}}\) is the target depth at which the reward saturates. This cap prevents SDR from rewarding arbitrarily many searches, while still favoring tasks whose easiest successful solution requires nontrivial search depth.

We now summarize the base Proposer reward as follows, where the first three cases correspond to the tool-use, format, and validity/parametric filters introduced above:
\begin{equation}
b_j^P =
\begin{cases}
-0.5, & \text{insufficient tool use},\\
0, & \text{incomplete task format},\\
0.1, & \text{invalid or tool-free answerable},\\
r_0, & C\in\{0,K_S\},\\
r_{\mathrm{sdr}}, & \text{otherwise}.
\end{cases}
\label{eq:base-rp}
\end{equation}

For the Solver, the base reward encourages answering correctly with tool-supported evidence:
\begin{equation}
b_i^S =
\begin{cases}
-0.5, & \text{no tool use},\\
0, & \text{invalid answer format},\\
z_i, & \text{otherwise}.
\end{cases}
\label{eq:base-rs}
\end{equation}

As training proceeds, the model can drift toward excessive \texttt{open} calls. Early in training, opening documents often provides useful information that helps the model answer correctly, so such behavior is reinforced by answer rewards. Over time, however, the policy may over-rely on \texttt{open}: many opens revisit already-seen documents rather than acquiring new evidence, producing long but shallow trajectories. A concrete over-opening case is provided in the Supplementary Material. Accordingly, we introduce the \textbf{Over-Opening Penalty (OOP)} to regularize this behavior in both roles, since the Proposer and Solver share the same policy \(\pi_\theta\). OOP penalizes the ratio of \texttt{open} to \texttt{search} actions rather than the raw number of opened documents, so it discourages redundant opening without suppressing legitimate multi-document exploration. Specifically, let \(n_{\mathrm{open}}(\tau)\) and \(n_{\mathrm{search}}(\tau)\) count \texttt{open} and \texttt{search} calls in a trajectory \(\tau\). For \(n_{\mathrm{search}}(\tau)>0\), define:
\begin{equation}
\rho_{\mathrm{oop}}(\tau)=
\mathrm{clip}\!\left(
\frac{n_{\mathrm{open}}(\tau)/n_{\mathrm{search}}(\tau)-\alpha_{\mathrm{oop}}}
{\beta_{\mathrm{oop}}-\alpha_{\mathrm{oop}}},
0,1
\right),
\label{eq:oop}
\end{equation}
where \(\alpha_{\mathrm{oop}}\) and \(\beta_{\mathrm{oop}}\) are the open-to-search ratios at which the penalty starts and reaches its maximum, so \(\rho_{\mathrm{oop}}\) rises linearly from \(0\) at \(\alpha_{\mathrm{oop}}\) to \(1\) at \(\beta_{\mathrm{oop}}\). The final rewards then subtract OOP with coefficient \(\lambda_{\mathrm{oop}}\):
\begin{equation}
r_j^P =
\begin{cases}
b_j^P - \lambda_{\mathrm{oop}}\rho_{\mathrm{oop}}(\tau_j^P), & b_j^P=r_{\mathrm{sdr}},\\
b_j^P, & \text{otherwise},
\end{cases}
\label{eq:rp}
\end{equation}
\begin{equation}
r_i^S =
\begin{cases}
b_i^S - \lambda_{\mathrm{oop}}\rho_{\mathrm{oop}}(\tau_i^S), &  b_i^S=z_i,\\
b_i^S, & \text{otherwise}.
\end{cases}
\label{eq:rs}
\end{equation}
\subsection{Training and Optimization}

After rewards are assigned, each retained seed document contributes two GRPO groups: a Proposer group containing the \(K_P\) task-generation rollouts, and a Solver group containing the \(K_S\) answer rollouts of the highest-reward passing task. Seeds whose highest task reward does not exceed \(r_0\) are discarded. Because of this seed-level gate, the final training data matches our high-quality search-data goal: retained tasks are grounded for cross-document retrieval, successful rollouts exhibit nontrivial search depth, and OOP regularizes tool use toward targeted search.

Within each retained group, rewards are normalized into advantages \(\hat{A}\) by subtracting the group mean and dividing by the group standard deviation with a small numerical constant. The same advantage is assigned to every generated token in the sample. We then update the shared policy with a token-level clipped objective and a KL regularizer,
\begin{equation}
\begin{aligned}
\mathcal{L}(\theta) ={}& -\,\mathbb{E}_{t}\!\left[
\min\!\big(\gamma_t\,\hat{A},\
\mathrm{clip}(\gamma_t,1-\epsilon_{\mathrm{low}},1+\epsilon_{\mathrm{high}})\,\hat{A}\big)
\right] \\
&+ \beta_{\mathrm{KL}}\,\mathbb{E}_{t}\!\left[
\mathrm{KL}\!\left(\pi_\theta\,\|\,\pi_{\mathrm{ref}}\right)
\right],
\end{aligned}
\label{eq:grpo}
\end{equation}
where the expectation is over unmasked generated tokens \(t\), \(\gamma_t=\pi_\theta(a_t\mid s_t)/\pi_{\theta_{\mathrm{old}}}(a_t\mid s_t)\) is the per-token importance ratio, and \((\epsilon_{\mathrm{low}},\epsilon_{\mathrm{high}})\) is the asymmetric clip range. Tool observations remain in the context but are masked out of the loss, so gradients flow only through model-generated tokens. Proposer and Solver samples are optimized jointly under the shared policy \(\pi_\theta\). Pseudocode for the full self-play iteration is provided in the Supplementary Material.

\begin{table*}[ht]
\centering
\small
\setlength{\tabcolsep}{4pt}
\begin{tabular}{lccccccc}
\toprule
Method & BrowseComp-Plus & BrowseComp & GAIA & SEAL-0 & WebWalkerQA & XBench & Avg. \\
\midrule
\multicolumn{8}{l}{\textit{Proprietary models and systems}} \\
OpenAI o3 \cite{openai2025o3} & 63.49 & 49.70 & 70.50 & 15.30 & 71.70 & 67.00 & -- \\
OpenAI o4-mini \cite{openai2025o3} & -- & 28.30 & 60.00 & -- & -- & -- & -- \\
GPT-5-high \cite{singh2025openai} & 70.12 & 54.90 & 76.40 & 43.20 & -- & 77.80 & -- \\
GPT-4.1 \cite{openai2025gpt41} & 35.42 & -- & -- & -- & -- & -- & -- \\
Claude Sonnet 4 \cite{anthropic2025claude4} & 36.75 & 12.20 & 68.30 & -- & 61.70 & 65.00 & -- \\
Claude Opus 4 \cite{anthropic2025claude4} & 36.14 & -- & -- & -- & -- & -- & -- \\
Gemini 2.5 Pro \cite{gemini2025gemini25} & 28.67 & -- & -- & -- & -- & -- & -- \\
Gemini 2.5 Flash \cite{gemini2025gemini25} & 33.01 & -- & -- & -- & -- & -- & -- \\
OpenAI DeepResearch \cite{openai2025deepresearch} & -- & 51.50 & 67.40 & -- & -- & -- & -- \\
\midrule
\multicolumn{8}{l}{\textit{Open-source models and systems}} \\
GLM-4.5 \cite{zeng2025glm} & -- & 26.40 & 66.00 & -- & 65.60 & 70.00 & -- \\
Kimi K2 \cite{team2025kimi} & -- & 14.10 & 57.70 & -- & 63.00 & 50.00 & -- \\
DeepSeek-V3.1 \cite{liu2024deepseek} & -- & 30.00 & 63.10 & -- & 61.20 & 71.00 & -- \\
Qwen3-32B \cite{yang2025qwen3} & 10.36 & -- & -- & -- & -- & -- & -- \\
SearchR1-32B \cite{jin2025search} & 10.36 & -- & -- & -- & -- & -- & -- \\
gpt-oss-120B-high \cite{agarwal2025gpt} & 42.89 & -- & -- & -- & -- & -- & -- \\
WebThinker-32B \cite{li2026webthinker}& -- & 2.80 & 48.50 & -- & 46.50 & 24.00 & -- \\
WebExplorer-8B \cite{liu2025webexplorer}& -- & 15.70 & 50.00 & -- & 62.70 & 53.70 & -- \\
WebDancer-QwQ \cite{wu2026webdancer} & -- & 3.80 & 51.50 & -- & 47.90 & -- & -- \\
WebShaper-72B \cite{tao2025webshaper}& -- & -- & 60.10 & -- & 52.20 & -- & -- \\
DeepDive-32B \cite{lu2025deepdive} & -- & 15.30 & -- & 25.50 & -- & 51.80 & -- \\
OffSeeker-8B \cite{zhou2026offseeker}& -- & 12.80 & 51.50 & -- & 61.70 & 49.00 & -- \\
WebSailor-72B \cite{li2025websailor}& -- & 12.00 & 55.40 & -- & -- & 55.00 & -- \\
WebSailor-V2-30B \cite{li2025websailorv2} & -- & 35.30 & 74.10 & -- & -- & 73.70 & -- \\
WebLeaper-C \cite{tao2025webleaper}& -- & 38.80 & 73.20 & 48.60 & -- & 72.00 & -- \\
BrowseMaster \cite{pang2025browsemaster} & -- & 30.00 & 68.00 & -- & 62.10 & 66.00 & -- \\
OpenResearcher-30B \cite{li2026openresearcher} & 54.80 & 26.30 & 64.10 & -- & -- & 65.00 & -- \\
REDSearcher-30B \cite{chu2026redsearcher} & -- & 42.10 & 80.10 & -- & -- & -- & -- \\
Tongyi-DR-30B \cite{team2025tongyi} & -- & 43.40 & 70.90 & -- & 72.20 & 75.00 & -- \\
MiroThinker-72B \cite{team2025mirothinker}& -- & 47.10 & 81.90 & 51.00 & 62.10 & 77.80 & -- \\
\midrule
\multicolumn{8}{l}{\textit{Ours (Qwen3.5 backbone)}} \\
Qwen3.5-9B \cite{qwen3.5} & 30.12 & 20.93 & 50.49 & 26.13 & 41.47 & 60.00 & 38.19 \\
SearchMaster-9B & 60.24 & 28.75 & 57.28 & 35.14 & 59.71 & 68.00 & 51.52 \\
Qwen3.5-27B \cite{qwen3.5} & 54.46 & 26.46 & 58.25 & 30.63 & 63.68 & 74.00 & 51.25 \\
SearchMaster-27B & 72.29 & 41.94 & 75.73 & 38.34 & 66.91 & 80.00 & 62.54 \\
\bottomrule
\end{tabular}
\caption{Main evaluation results. All columns report accuracy (\%), and Avg.\ is the arithmetic mean over the six benchmarks. External baselines are reported from their source papers when available.}
\label{tab:main-results}
\end{table*}

\section{Experiments}

\subsection{Experimental Setup}

\textbf{Implementation Details.}
The local search database \(\mathcal{D}\) is the offline corpus released by OpenResearcher \cite{li2026openresearcher} (about 15M documents, \(\sim\)11B tokens), indexed with a Qwen3-Embedding-8B \cite{yang2025qwen3} FAISS retriever. We experiment with both Qwen3.5-9B and Qwen3.5-27B \cite{qwen3.5} backbones. We train with GRPO at a constant learning rate of \(1\times10^{-6}\), clip range \((\epsilon_{\mathrm{low}},\epsilon_{\mathrm{high}})=(0.2,0.28)\), and KL weight \(\beta_{\mathrm{KL}}=0.001\). Each iteration randomly samples seed documents from \(\mathcal{D}\) and, after the four filters and the seed-level quality gate, retains 64 qualified seed documents for training. For each seed document, the Proposer samples \(K_P=8\) rollouts, and each passing candidate launches \(K_S=8\) independent Solver rollouts for verification and reward computation, with up to 200 tool calls per rollout. Although Solver rollouts are generated for all passing candidates to compute rewards, only the Solver group of the highest-reward passing task is used for optimization, so each retained seed contributes \(K_P+K_S\) training samples. Each iteration therefore contributes a global optimization batch size of \(64\times(K_P+K_S)=1{,}024\). For reward design, we set \(r_0=0.2\), \(H_{\mathrm{sdr}}=10\), and OOP parameters \(\alpha_{\mathrm{oop}}=1.5\), \(\beta_{\mathrm{oop}}=2.5\), and \(\lambda_{\mathrm{oop}}=0.5\). Rollouts use a 256K-token context and decoding temperature \(1.0\). We train for 20 iterations, so SearchMaster learns from \(20\times64=1{,}280\) qualified seed documents in total. Prompt templates, filter thresholds, and additional training hyperparameters are provided in the Supplementary Material.

\textbf{Benchmarks.}
We evaluate SearchMaster on both offline and online search benchmarks. The offline benchmark is BrowseComp-Plus \cite{chen2026browsecomp}, with 830 browsing questions over an offline retrieval corpus drawn from a different source than our OpenResearcher training corpus. For transfer to live web search, we use five online search benchmarks: BrowseComp \cite{wei2025browsecomp} (1{,}266 hard-to-find factual questions), GAIA \cite{mialon2024gaia} (103 text-only tool-use assistant tasks), SEAL-0 \cite{pham2025sealqa} (111 search-intensive long-tail questions), WebWalkerQA \cite{wu2025webwalker} (680 multi-page navigation questions), and XBench-DeepSearch \cite{chen2025xbench} (100 web-search questions). During evaluation, we do not use any context-management method; the context length is 256K tokens, decoding temperature is \(1.0\), and outputs are scored by GPT-5 \cite{singh2025openai} against the ground-truth answers, with the judging prompt provided in the Supplementary Material.

\subsection{Main Results}

Table~\ref{tab:main-results} reports the main results. On BrowseComp-Plus, which supports multiple retrievers, we report all methods under its Qwen3-Embedding-8B \cite{yang2025qwen3} retriever setting for fair comparison.

\textbf{Self-play yields substantial gains on BrowseComp-Plus.}
SearchMaster raises the Qwen3.5-9B backbone from 30.12\% to 60.24\%, an absolute gain of 30.1 points. This lifts the 9B model above much larger open-source systems such as gpt-oss-120B-high (42.89\%) and proprietary models such as GPT-4.1 (35.42\%) and Claude Opus 4 (36.14\%), while narrowing the gap to stronger reasoning systems such as OpenAI o3 (63.49\%). Notably, this is achieved purely through self-play over a local search environment, without human-written questions or advanced teacher models. These results show that a single model can bootstrap strong deep-search ability entirely from its own generated tasks. The Supplementary Material examines the source of this gain: the base model already issues valid tool calls but searches ineffectively, whereas SearchMaster learns to search more deeply and follow evidence across documents.

\textbf{The gains persist at a larger model scale.}
Starting from the substantially stronger Qwen3.5-27B backbone, SearchMaster improves BrowseComp-Plus from 54.46\% to 72.29\% (\(+17.8\) points) and average accuracy from 51.25\% to 62.54\% (\(+11.3\) points). SearchMaster-27B outperforms its base model on every benchmark in Table~\ref{tab:main-results}, showing that self-play remains effective even when the initial policy already has considerably stronger search capability. Together with the 9B results, this suggests that SearchMaster improves search behavior rather than merely compensating for the weaknesses of a small backbone.

\textbf{The learned search behavior generalizes to live web search.}
Although SearchMaster is trained entirely in an offline search environment, both SearchMaster-9B and SearchMaster-27B improve their corresponding backbones on all five online benchmarks that query the live, open web. This consistent transfer from offline training to live web search shows that SearchMaster equips models with search behavior that generalizes beyond their training environment.

\subsection{Ablation Study}

In this section, we study the contribution of ECG, SDR, and OOP using the default Qwen3.5-9B backbone. All variants use the same training budget and filtering pipeline unless otherwise noted. We first train a naive self-play baseline that keeps the same task filters and Solver verification as SearchMaster, but uses rollout accuracy as the task-difficulty signal, via a success-rate term \(1 - \left| 2C/K_S - 1 \right|\), where \(C\) is the number of successful Solver rollouts among \(K_S\) attempts, so that the reward peaks when half of the rollouts succeed. Starting from this baseline, ECG changes the Proposer prompt into evidence-chain construction, SDR replaces the success-rate difficulty signal with the search-depth reward, and OOP adds the over-opening penalty.

Table~\ref{tab:training-variants} reports the results on BrowseComp-Plus. Naive self-play alone lifts the backbone from 30.12\% to 45.18\%, showing that self-play over a local corpus already provides a useful signal. Adding ECG, SDR, and OOP improves accuracy monotonically, and the full SearchMaster reaches 60.24\%. These mechanisms improve complementary parts of the self-play loop: chain-based task construction (ECG), search-depth-aware task selection (SDR), and tool-use regularization (OOP). The rest of this section examines whether each mechanism produces its intended behavioral effect, beyond the final accuracy.

\begin{table}[t]
\centering
\small
\setlength{\tabcolsep}{4pt}
\begin{tabular}{l|ccc|c}
\toprule
Variant & ECG & SDR & OOP & BrowseComp-Plus \\
\midrule
Qwen3.5-9B & - & - & - & 30.12 \\ \hline
naive self-play & & & & 45.18 \\
\quad + ECG & \checkmark & & & 53.25 \\
\quad + SDR & & \checkmark & & 52.89 \\
\quad + ECG + SDR & \checkmark & \checkmark & & 57.71 \\
SearchMaster & \checkmark & \checkmark & \checkmark & \textbf{60.24} \\
\bottomrule
\end{tabular}
\caption{Ablation on BrowseComp-Plus. ECG, SDR, and OOP each contribute a further gain over naive self-play.}
\label{tab:training-variants}
\end{table}

\noindent\textbf{Does ECG Reduce Pseudo Multi-Hop Tasks?}
ECG makes evidence-chain construction explicit before question generation, so we directly test whether it reduces pseudo multi-hop tasks. We compare three settings on the same 1{,}000 seed documents: Qwen3.5-9B with the naive self-play prompt, Qwen3.5-9B with the ECG prompt, and the trained SearchMaster with ECG. Each setting generates 10 tasks per seed document. A GLM-5 judge labels each task as \textit{True Multi-Hop} if it is answerable only by combining evidence across several documents, \textit{Pseudo Multi-Hop} if it superficially involves multiple entities but is solvable by a single lookup, and \textit{Invalid} if it leaks the answer or is unsupported by the retrieved evidence.

Table~\ref{tab:task-quality} shows that adding ECG to the base model increases true multi-hop tasks from 24.2\% to 45.6\% and reduces invalid tasks from 28.7\% to 15.8\%. After SearchMaster training, the true multi-hop rate further rises to 78.6\%, while pseudo multi-hop drops to 15.0\%. The Supplementary Material further compares Qwen3.5-9B with the naive and ECG prompts on the same seeds. In one case, the naive prompt searches around seed-level facts and asks a seed-answerable question; in another, it reaches related documents but leaves their links unused in the final question. These examples suggest that ECG helps in two ways: it drives the model to explore beyond seed-level facts, and it requires the explored evidence to be organized into an explicit chain before question generation. Together, these effects turn loosely related browsing into task-relevant cross-document dependencies.

\begin{table}[t]
\centering
\small
\setlength{\tabcolsep}{3pt}
\begin{tabular}{lccc}
\toprule
Method & True Multi-Hop $\uparrow$ & Pseudo $\downarrow$ & Invalid $\downarrow$ \\
\midrule
Qwen3.5-9B + naive & 24.2 & 47.1 & 28.7 \\
Qwen3.5-9B + ECG & 45.6 & 38.6 & 15.8 \\
SearchMaster + ECG & \textbf{78.6} & \textbf{15.0} & \textbf{6.4} \\
\bottomrule
\end{tabular}
\caption{Task-quality comparison across three settings. Each setting generates 10 tasks for each of the same 1{,}000 seed documents (10{,}000 per setting), judged by GLM-5 \cite{zeng2026glm}.}
\label{tab:task-quality}
\end{table}

\begin{figure}[t]
  \centering
  \includegraphics[width=\linewidth]{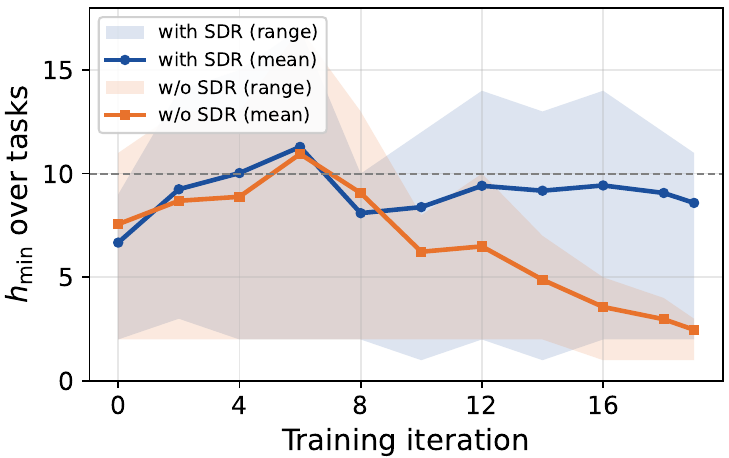}
  \caption{Search depth \(h_{\min}\) of the training tasks over self-play. Lines show the mean and bands show the range across tasks.}
  \label{fig:sdr-hmin}
\end{figure}

\noindent\textbf{Does SDR Keep the Training Tasks Search-Intensive?}
SDR scores a task by \(h_{\min}\), the minimum search depth among successful Solver rollouts (Eq.~\ref{eq:sdr}). Because \(h_{\min}\) reflects the easiest successful solution, a large value indicates that the task cannot be solved by a shallow lookup. To test whether SDR keeps retained tasks search-intensive during training, we compare the naive self-play baseline, which uses a success-rate difficulty signal, with the \(+\)SDR setting from Table~\ref{tab:training-variants}. We track \(h_{\min}\) across iterations in Figure~\ref{fig:sdr-hmin}, discarding a few tasks with extremely large \(h_{\min}\) for visualization.

Early in training, both settings produce tasks with increasing \(h_{\min}\), suggesting that self-play initially discovers tasks requiring more search. Later, the two runs diverge. Under the success-rate signal, the Proposer has no direct incentive to preserve search depth: as the Solver improves, retained tasks become solvable with fewer searches, and \(h_{\min}\) settles around \(2\)--\(3\). SDR instead rewards tasks whose easiest successful solution still requires substantial search, so \(h_{\min}\) remains in the \(8\)--\(10\) range even as the Solver strengthens. This shows that SDR prevents the training tasks from decaying into shallow lookup tasks and maintains a multi-step search signal.

\begin{figure}[t]
  \centering
  \includegraphics[width=\linewidth]{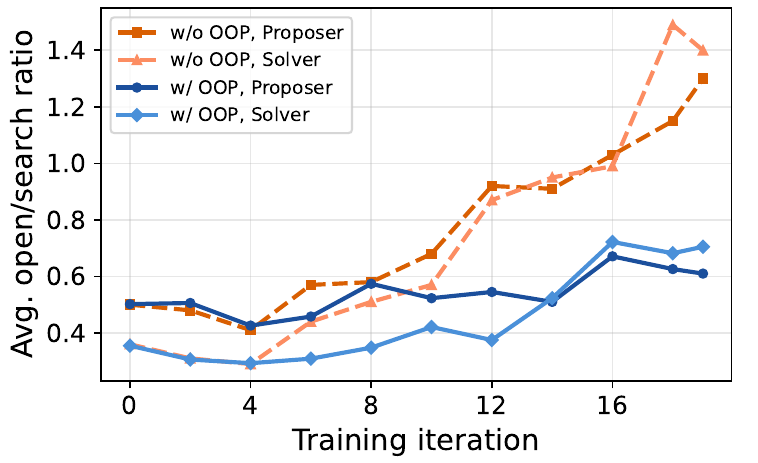}
  \caption{Average open/search ratio of the Proposer and the Solver over training. Without OOP, both roles drift toward higher open/search ratios; with OOP, the average ratio stays low for both roles.}
  \label{fig:oop-ratio}
\end{figure}

\noindent\textbf{Does OOP Control Open Overuse?}
To evaluate whether OOP regulates tool use, we compare two settings from Table~\ref{tab:training-variants}: \(+\)ECG\(+\)SDR without OOP and the full SearchMaster with OOP. Since the Proposer and Solver share the same policy, we track the average open-to-search ratio of both roles across training iterations (Figure~\ref{fig:oop-ratio}).

Without OOP, the open-to-search ratio of both roles increases steadily, indicating a drift toward opening documents rather than issuing targeted searches. To inspect this behavior, the Supplementary Material analyzes late-training Solver rollouts with at least twice as many opens as searches. In these over-opening rollouts, 56.3\% of the extra \texttt{open} calls simply re-open already-seen documents rather than acquiring new evidence. With OOP, the ratio of both roles remains low, showing that the penalty curbs over-opening while still allowing the searches and opens needed for genuine multi-document tasks.

\section{Conclusion}

We introduce SearchMaster, a self-play framework for training search agents from tasks discovered in a local search environment. The central challenge is not merely question generation, but transforming autonomous exploration into useful training data for search behavior. SearchMaster addresses this challenge through ECG for evidence-grounded task construction, SDR for search-depth difficulty calibration, and OOP for preventing over-opening in tool-use trajectories. Across six deep-search benchmarks, SearchMaster raises average accuracy from 38.19\% to 51.52\% with Qwen3.5-9B and from 51.25\% to 62.54\% with Qwen3.5-27B. The corresponding BrowseComp-Plus gains are 30.1 and 17.8 points, demonstrating that grounded and regulated self-play remains effective across model scales.

While SearchMaster reduces dependence on annotated QA data, it does not eliminate all costs. Training still requires a searchable environment, multiple Solver rollouts per proposed task, and Verifier calls for grounding, parametric filtering, and correctness. The current implementation operates in a local search environment, which improves reproducibility, though it may limit coverage of the open web's full diversity and volatility. Future work can extend SearchMaster beyond local corpora to more diverse and dynamic web environments, while reducing the rollout and verification cost of self-play training.

\bibliography{aaai2027}

\end{document}